\documentclass[runningheads]{llncs}
\usepackage{authblk}
\usepackage{graphicx}
\usepackage{amsmath,amssymb} % define this before the line numbering.
\usepackage{color}
\usepackage{booktabs}
\usepackage{multirow}
\usepackage{epsfig}
\usepackage[accsupp]{axessibility}  % Improves PDF readability for those with disabilities.
\usepackage{caption}
\usepackage{subcaption}
\usepackage{hyperref}

\def\ie{\emph{i.e. }}

\begin{document}
\pagestyle{headings}
\mainmatter

\def\ACCV22SubNumber{125}  % Insert your submission number here

%===========================================================
\title{GaitStrip: Gait Recognition via Effective Strip-based Feature Representations and Multi-Level Framework} % Replace with your title
\titlerunning{GaitStrip}
\authorrunning{M. Wang et al.}

%\author{Anonymous ACCV 2022 submission}
\author{Ming Wang\inst{1}  \and
Beibei Lin\inst{2}\and
Xianda Guo\inst{3} \and
Lincheng Li\inst{4}\and
Zheng Zhu\inst{3}\and 
\\Jiande Sun\inst{5}\and
Shunli Zhang\inst{1}\thanks{Shunli Zhang is the corresponding author. slzhang@bjtu.edu.cn}\and
Yu Liu\inst{1}\and
Xin Yu\inst{6}}

%  slzhang@bjtu.edu.cn
% \institute{\inst{1}Beijing Jiaotong University,\inst{2} National University of Singapore, \\
% \inst{3}PhiGent Robotics, \inst{4}NetEase Fuxi AI Lab, \inst{5}Tsinghua University, \\ \inst{6}Shandong Normal University,\inst{7}University of Technology Sydney \\
% }

\institute{\textsuperscript{1}Beijing Jiaotong University,\inst{2} National University of Singapore, \\
\inst{3}PhiGent Robotics, \inst{4}NetEase Fuxi AI Lab,  \\ \inst{5}Shandong Normal University,\inst{6}University of Technology Sydney \\
}
\maketitle

%===========================================================
\begin{abstract}
Many gait recognition methods first partition the human gait into N-parts and then combine them to establish part-based feature representations.
Their gait recognition performance is often affected by partitioning strategies, which are empirically chosen in different datasets. However, we observe that strips as the basic component of parts are agnostic against different partitioning strategies. Motivated by this observation, we present a strip-based multi-level gait recognition network, named GaitStrip, to extract comprehensive gait information at different levels.
To be specific, our high-level branch explores the context of gait sequences and our low-level one focuses on detailed posture changes.
We introduce a novel StriP-Based feature extractor (SPB) to learn the strip-based feature representations by directly taking each strip of the human body as the basic unit. 
Moreover, we propose a novel multi-branch structure, called Enhanced Convolution Module (ECM), to extract different representations of gaits. ECM consists of the Spatial-Temporal feature extractor (ST), the Frame-Level feature extractor (FL) and SPB, and has two obvious advantages: 
First, each branch focuses on a specific representation, which can be used to improve the robustness of the network. Specifically, ST aims to extract spatial-temporal features of gait sequences, while FL is used to generate the feature representation of each frame.
Second, the parameters of the ECM can be reduced in test by introducing a structural re-parameterization technique.
Extensive experimental results demonstrate that our GaitStrip achieves state-of-the-art performance in both normal walking and complex conditions. The source code is published at \url{https://github.com/M-Candy77/GaitStrip}. 

\end{abstract}

\begin{figure}[t]
\centering
\includegraphics[width=0.7\textwidth]{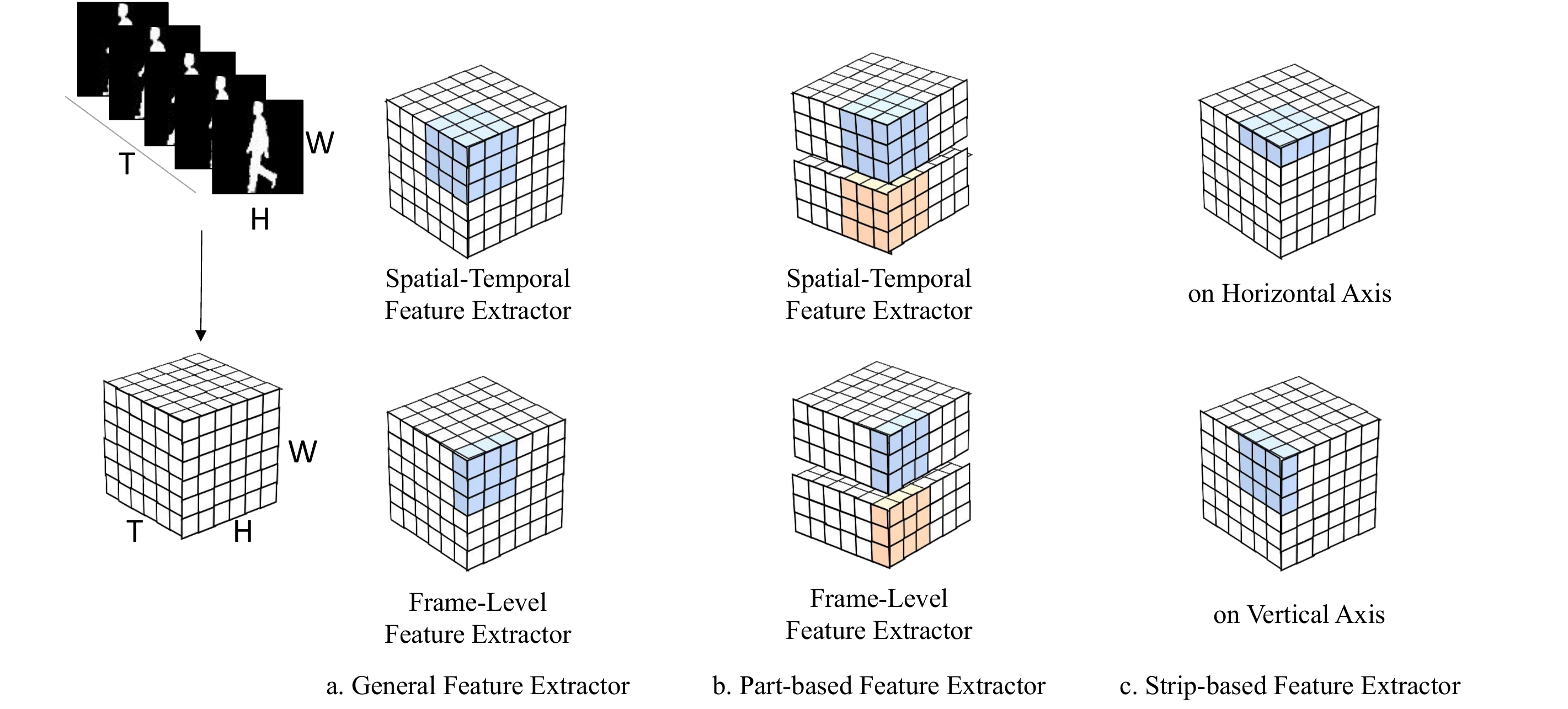}
%\vspace*{-1em}
\caption{Visualization of feature extractors of different methods.}
% \vspace*{-2em}
\label{fig_v}
\end{figure}

%===========================================================
\section{Introduction}

\label{sec:intro}
Gait recognition is one of the most popular biometric techniques. Since it can be used in a long-distance condition and cannot be disguised, gait recognition is widely applied in video surveillance and access control systems. However, this technology has experienced a huge challenge due to the complexity of the external environment, such as cross-view, speed changes, bad weathers and variations in appearances~\cite{yeoh2016clothing,yu2006framework,connor2018biometric,jin2022unsupervised,jin2021dc}.

Recently, many Convolutional Neural Networks (CNNs) based gait recognition frameworks have been proposed to generate discriminative feature representations
~\cite{shiraga2016geinet,chao2019gaitset,fan2020gaitpart,zhang2016siamese,zhang2019cross,wu2020condition,li2020end,chai2021silhouette,lin2021multi,song2019gaitnet,li2020gait,li2019attentive,li2020gait1,yu2021hid,zhu2021gait,lin2021gaitmask,lin2022gaitgl,shen2022gait,huang2022enhanced}. 
As shown in Fig.\ref{fig_v} (a), some researchers extract gait features directly from the whole gait sequence, which captures global context information of gait sequences ~\cite{shiraga2016geinet,chao2019gaitset,zhang2016siamese}. As those methods take the human gait as a unit to extract features, some local gait changes that are important for gait recognition might not be fully captured, which may affect the recognition performance. 
On the other hand, some other researchers~\cite{fan2020gaitpart,zhang2019cross} propose part-based feature representation to represent the human gait, which is shown in Fig.\ref{fig_v} (b). They first partition the human gait into N-parts and then extract the detailed information of each part. 
Although carefully choosing the number of partitions in different convolutional layers can achieve appealing performance, it is unclear how to build an accurate part-based model on new datasets, which limits the generalization of the methods.

According to these findings, we argue that the part-based feature representation is not a general feature representation for gait recognition. Hence, we question \textit{whether there is a gait descriptor that is insensitive to various partitions?} Through carefully analysis of recent part-based methods, we find that strips are the minimal effective representation elements for gaits instead of parts. Using strips, we will be able to circumvent the handcrafted partition in part-based methods. 
% \textcolor{red}{Hence, we propose the strip-based partition which does not require empirical partitioning is more general.} 
As shown in Fig.\ref{fig_v} (c), the strip can be considered as an extreme form of the part-based representation, thus it is not necessary to manually determine the reasonable number of the parts.
Motivated by this observation, we propose a new gait recognition network, called GaitStrip, to learn more discriminative feature representations based on strips. 
Specifically, GaitStrip is implemented under a multi-level framework to improve the representation capability. The multi-level framework includes two branches, i.e., the low-level branch and the high-level one. In particular, the high-level branch extracts the global context information from low-resolution gait images, while the low-level one captures more details from high-resolution images.

Furthermore, we introduce Enhanced Convolution Module (ECM), as a multi-branch block, to our GaitStrip. 
% Furthermore, GaitStrip is built by a new multi-branch block, dubbed Enhanced Convolution Module (ECM), to effectively utilize different representations at different branches. 
ECM includes three branches, i.e., the StriP-Based feature extractor (SPB), the Spatial-Temporal feature extractor (ST) and the Frame-Level feature extractor (FL), where each branch corresponds to a specific representation. Specifically, SPB is designed to generate strip-based feature representations by taking each strip of the human body as a basic unit, ST aims to extract spatial-temporal information of a gait sequence, and FL is used to extract each frame's spatial features.
On the other hand, we introduce a structural re-parameterization technique to reduce the parameters of the ECM module in test~\cite{ding2021repvgg}. Specifically, the parameters of SPB, ST and FL can be merged into a single $3\times 3\times 3$ convolution.

After feature extraction, we obtain an effective feature representation by using temporal aggregation and spatial mapping operations. The temporal aggregation ensembles temporal information of a variable-length gait sequence~\cite{lin2020gait}. The spatial mapping first partitions the feature maps into multiple horizontal vectors and aggregates each vector by Generalized-Mean (GeM) pooling operations~\cite{radenovic2018fine} for better representation. 
Extensive experiments on widely-used gait recognition benchmarks demonstrate that our GaitStrip outperforms the state-of-the-arts significantly.

The main contributions of the proposed method are three-fold, shown as follows:
\begin{itemize}
\item 

Based on the observation that the strip-based method can achieve more effective gait representations than part-based partitioning,
% We observe that compared with recent part-based partitions, the strip-based partition can eliminate the effect of partitioning and achieve more effective gait representations. 
we propose a multi-level gait recognition framework with strip to extract more comprehensive gait features, in which the high-level representation contains the context information while the low-level representation extracts local details of gait sequences.
\item We develop an effective enhanced convolution module including three branches, which can not only take the advantage of both frame-level and spatial-temporal features but also use SPB to enhance the representation ability. Furthermore, we use the structural re-parameterization technique to reduce the parameters for high efficiency in test.
\item We compare the proposed method with several state-of-the-art methods on two public datasets, CASIA-B and OUMVLP. The experimental results demonstrate that the performance of the proposed method achieves superior performance to these approaches.
\end{itemize}

%===========================================================
\section{Related Work}

\subsection{Gait Recognition}

Existing gait recognition methods can be divided into two types, \ie, global-based and local-based.

The global-based methods usually take the human gait as a sample to generate global feature representations~\cite{shiraga2016geinet,zhang2016siamese,wu2016comprehensive}.
%{\color{red}which often pay attention to movement changes of the whole body not some essential body parts.
%The gait templates aggregate periodic gait information, which include many temporal templates.}
For instance, Shiraga et al.~\cite{shiraga2016geinet} first calculate the Gait Energy Image (GEI) by using the mean function to compress the temporal information of gait sequences, and then utilize 2D CNNs to extract gait features. 
% Zhang et al. \cite{zhang2016siamese} propose a siamese framework based on 2D CNNs to identify Human IDs by taking advantage of the GEI. Wu et al.\cite{wu2016comprehensive} propose three different frameworks with 2D CNNs to generate feature representations from the GEI. 
However, the generation of the GEI causes the loss of temporal information, which may degrade the representation ability. Thus, some other researchers ~\cite{chao2019gaitset,hou2021set,chao2021gaitset,zhang2019comprehensive} use 2D CNNs to extract each frame's feature before building the template. On the other hand, some researchers ~\cite{thapar2019gait,wolf2016multi,lin2020gait}  extract spatial-temporal information from gait sequences for representation.
Recently, 3D CNN has been used in gait recognition to learn the spatial-temporal representation of the entire gait sequence.
For example, Lin et al.~\cite{lin2020gait} use 3D CNNs to extract spatial-temporal information, and employ temporal aggregation to integrate temporal information, addressing the variable-length issue of video sequences.

The local-based methods usually take the part of the human gait as input to establish the part-based feature representations~\cite{fan2020gaitpart,zhang2019cross}. 
%{\color{red}The different parts often process different information on human movement.} 
For example, Fan et al.~\cite{fan2020gaitpart} propose a focal convolution layer to extract part-based gait features. The focal convolutional layer first splits the feature maps into several local parts and then uses a shared convolution to extract each part's feature. Zhang et al.~\cite{zhang2019cross} first partition the human gait into four parts and then use 2D CNN to obtain feature representations of each part.
However, these local-based methods need to predefine the number of partitions for specific datasets, which limit the generalization ability.

\subsection{Strip-based Modeling}
\label{related_sbm}
Recently many strip-based modeling methods have been proposed in the visual field. For example, Ding et al.~\cite{ding2019acnet} propose a novel block, called Asymmetric
Convolution Block (ACB), to generate discriminative feature representations. They use 1D asymmetric forms (e.g. $3\times1$ Conv and $1\times3$ Conv) to improve the feature representation ability of the standard square-kernel convolution ($3\times3$ Conv). Note that the asymmetric convolutions can exploit the information of the horizontal and vertical strips. In particular, the asymmetric convolutions can be fused into the original square-kernel convolution. 
Huang et al.~\cite{huang2019ccnet} propose the CCNet network to capture global contextual information. CCNet which is built with Criss-Cross Attention blocks models the relationships of horizontal and vertical strips. 

However, the aforementioned methods only focus on the spatial strip-based information, which do not capture the temporal changes of each strip. Therefore, in this paper, we propose a novel strip-based feature extractor, which can be used to establish each strip's spatial-temporal information.
In particular, as far as we know, GaitStrip is the first network which models strip-based feature representations in gait recognition.

\section{Proposed Method}
In this section, we first overview the whole gait recognition framework. Then, we describe the enhanced convolution module, the multiple-level structure and feature mapping in detail. Finally, we introduce the strategies of training and test. 

\begin{figure*}[t]
\centering
%\vspace*{-2em}
\includegraphics[width=1\textwidth]{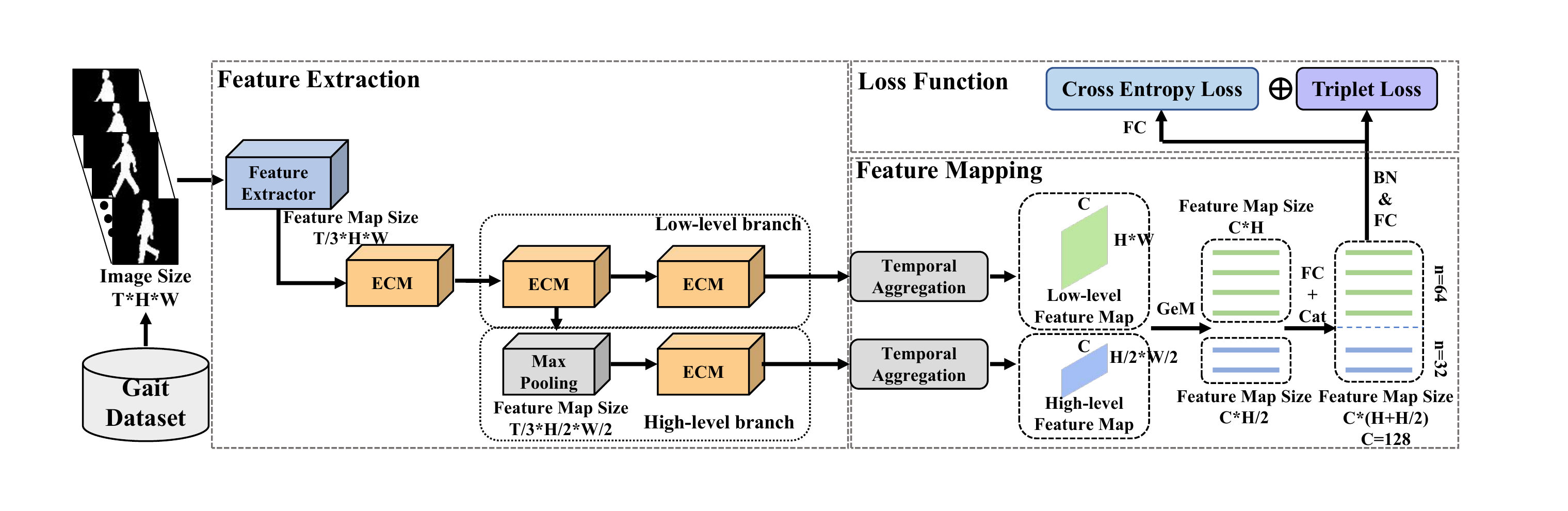}
%\vspace*{-3em}
\caption{Overview of the entire gait recognition framework.}
% \vspace*{-2em}
\label{overview}
\end{figure*}

\subsection{Overview}
\label{hg_level}
The proposed gait recognition framework, GaitStrip, which includes the feature extraction stage and feature mapping stage is shown in Fig.~\ref{overview}. The GaitStrip is constructed based on 3D convolutions, which can effectively extract spatial-temporal information of gait sequences.
During the feature extraction stage, a novel enhanced convolution module which uses both frame-level feature extractor and strip-based feature extractor to improve the representation ability of the traditional spatial-temporal feature extractor is proposed. 
%Specifically, FL aims to generate frame-level feature representation and SPB establishes each strip's spatial-temporal feature representation on the horizontal and vertical axis.
Then, we design the multi-level framework which includes both the high-level and the low-level branches.
%Then, the multi-level structure is designed to extract both high-level and low-level features. The low-level branch directly extracts features from the large-size feature maps, which focuses on details of the human body. By contrast, the high-level one works on down-sampled feature maps which extract more abstract information. 
During the feature mapping stage, the temporal aggregation operation is introduced to integrate the temporal information of feature maps~\cite{lin2020gait}. Then, the feature maps are partitioned into multiple horizontal vectors and the information is aggregated by Generalized-Mean (GeM) pooling~\cite{radenovic2018fine}. Finally, a combined loss function consisting of both cross-entropy loss and triplet loss is employed to train the proposed network.

\subsection{Enhanced Convolution Module}
Recently, many excellent feature extractors have been proposed to extract robust gait features, which can be divided into two types. One is the frame-level feature extractor which extracts gait features of each frame~\cite{chao2021gaitset,chao2019gaitset,fan2020gaitpart}, and the other one is the spatial-temporal feature extractor which generates spatial-temporal feature representations of a gait sequence~\cite{thapar2019gait,wolf2016multi,lin2020gait}.

Assume that the feature map $X_{in}\in \mathbb{R}^{C_{in} \times T_{in} \times H_{in} \times W_{in}}$ is the input of a convolution operation, where $C_{in}$ is the number of channels, $T_{in}$ is the length of gait sequences and ($H_{in}$,$W_{in}$) is the image size of each frame. 
The frame-level and spatial-temporal feature extractors can be designed as
\begin{equation}\label{FL}
X_{FL} = c^{1 \times 3 \times 3}(X_{in}),
\end{equation}
\begin{equation}\label{ST}
X_{ST} = c^{3 \times 3 \times 3}(X_{in}),
\end{equation}
where $c^{a \times b \times c}(\cdot)$ represents the 3D convolution with  kernel size $(a, b, c)$. $X_{FL}\in \mathbb{R}^{C_{out} \times T_{in} \times H_{in} \times W_{in}}$ and $X_{ST}\in \mathbb{R}^{C_{out} \times T_{in} \times H_{in} \times W_{in}}$ are the output of the frame-level and spatial-temporal feature extractors, respectively.

The frame-level features ignore the temporal information of the gait sequence, while the spatial-temporal features focus on the spatial-temporal changes, which may not pay enough attention to the detailed information of each frame. Thus, we propose a combined framework which takes advantage of frame-level and spatial-temporal information as our backbone. The combined structure includes two branches, i.e. the spatial-temporal feature extractor and frame-level feature extractor, which can be designed as
%As shown in Fig.\ref{ECM_fig}(c), the combined structure includes two branches, i.e. the spatial-temporal feature extractor and frame-level feature extractor, which can be designed as
\begin{equation}
X_{STFL} = X_{FL} + X_{ST}.
\end{equation}

%Although these methods are widely used, they do not pay enough attention to the part-based information of gait sequences in the feature extraction stage. In the previous literature, Fan et al. \cite{fan2020gaitpart} first split the human body into several parts in the horizontal axis and then use a shared convolution to extract each part's features. 
%However, the number of parts needs to select empirically in different datasets, which is out of the justified guidance.
To further improve the global representation and address the inflexibility issue in the part-based representation, we present a StriP-Based feature extractor (SPB) which extracts strip-based features on horizontal axis and vertical axis, respectively. The strip-based feature extractor on horizontal axis first splits the human body into multiple horizontal strips and then applies convolution to extract spatial-temporal information of each horizontal strip. This extractor can be denoted as
\begin{equation}\label{HSPB}
X_{SPB-H} = c^{3 \times 1 \times 3}(X_{in}),
\end{equation}
where $c^{3 \times 1 \times 3}(\cdot)$ denotes the 3D convolution with kernel size (3, 1, 3). $X_{SPB-H} \in \mathbb{R}^{C_{out} \times T_{in} \times H_{in} \times W_{in}}$ is the output of this extractor.

Similarly, the strip-based feature extractor is used for the vertical strip's spatial-temporal extraction, represented as
\begin{equation}\label{VSPB}
X_{SPB-V} = c^{3 \times 3 \times 1}(X_{in}),
\end{equation}
where $c^{3 \times 3 \times 1}(\cdot)$ denotes the 3D convolution with kernel size (3, 3, 1). $X_{SPB-V} \in \mathbb{R}^{C_{out} \times T_{in} \times H_{in} \times W_{in}}$ is the output of this extractor.
Finally, by combining the horizontal-based and vertical-based feature extractors, the strip-based feature extractor can obtain the following feature maps
\begin{equation}\label{equ_spb}
X_{SPB} = X_{SPB-H} + X_{SPB-V}.
\end{equation}

\begin{figure*}[ht]
\centering
%\scalebox{0.9}[0.75]{\includegraphics[width=\textwidth]{Figure3.pdf}}
\includegraphics[width=0.9\textwidth]{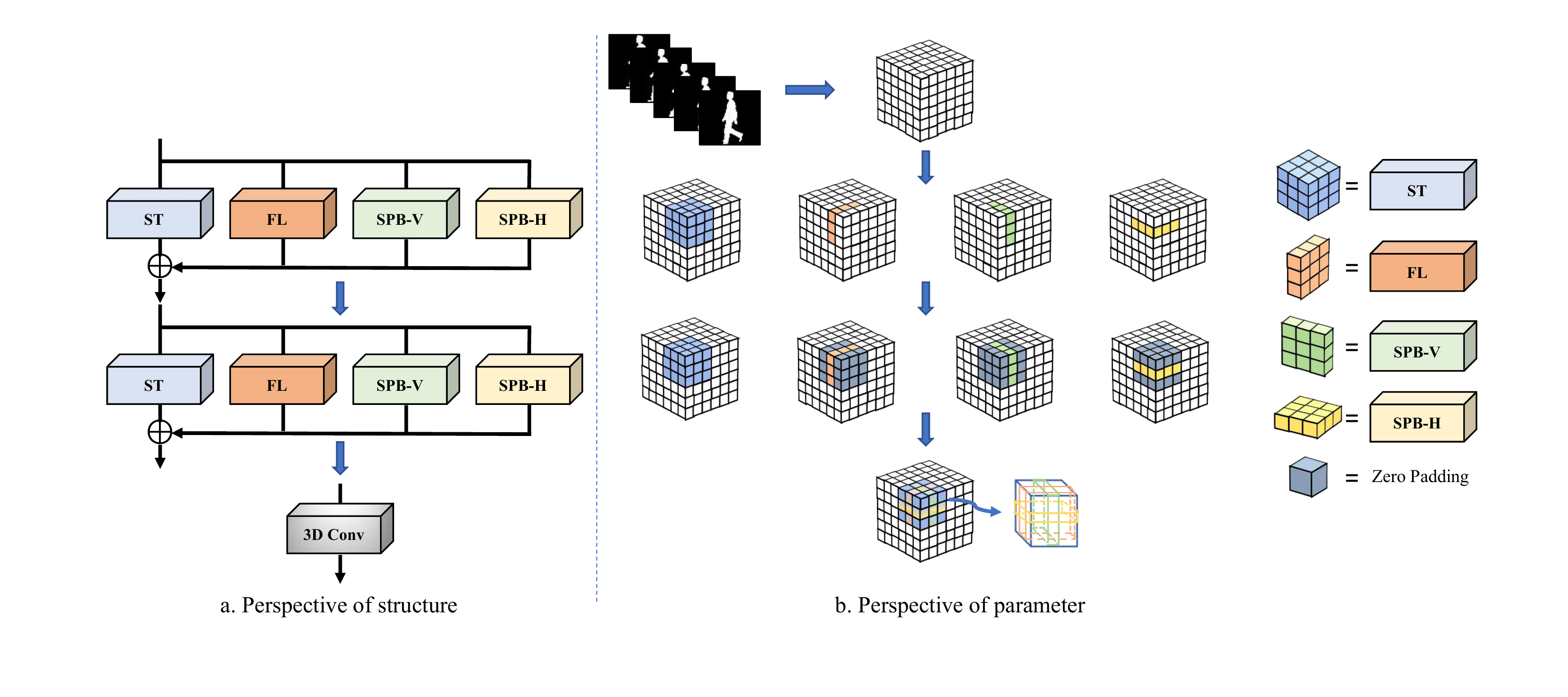}
%\vspace{-3em}
%\caption{Overview of the enhanced convolution module.}
\caption{Overview of the enhanced convolution module. ST represents the Spatial-Temporal feature extractor, FL represents the Frame-Level feature extractor, SPB-V represents StriP-Based feature extractor in vertical and SPB-H represents StriP-Based feature extractor in horizontal. }
% \vspace*{-2em}
\label{ECM_fig}
\end{figure*}

The proposed SPB can be used to enhance the feature representation ability of the traditional feature extractor. 
By combining SPB with aforementioned spatial-temporal feature extractor and frame-level feature extractor, as shown in Fig.~\ref{ECM_fig}, the ECM module can be obtained as follows
% Thereby, we propose a novel convolution modules, called ECM, 
\begin{equation}\label{equ_ECM}
X_{ECM} = X_{ST} + X_{FL} + X_{SPB}.
\end{equation}
Thus more comprehensive feature representations can be achieved.

\subsection{Structural Re-Parameterization}
To reduce the parameters of the proposed ECM, we introduce the structural re-parameterization \cite{ding2021repvgg} method to ensemble different convolutions during the test stage. As shown in Equ.\ref{equ_ECM}, the ECM block includes four convolutions: $c^{3 \times 3\times 3}(\cdot)$, $c^{1 \times 3\times 3}(\cdot)$, $c^{3 \times 1\times 3}(\cdot)$ and $c^{3 \times 3\times 1}(\cdot)$. During the test stage, these convolutions can be integrated into a single 3D convolution $c_{emb}^{3 \times 3\times 3}(\cdot)$, which can be designed as
\begin{equation}\label{c_emb}
c_{emb}^{3 \times 3\times 3} = c^{3 \times 3\times 3} + c_t^{3 \times 3\times 3} + c_h^{3 \times 3\times 3} + c_w^{3 \times 3\times 3},
\end{equation}
where $c_t^{3 \times 3\times 3}$, $c_h^{3 \times 3\times 3}$ and $c_w^{3 \times 3\times 3}$ are zero-padding expansions of $c^{1 \times 3\times 3}$, $c^{3 \times 1\times 3}$ and $c^{3 \times 3\times 1}$, respectively, to make the kernels maintain the same dimensions. 
According to Equ.\ref{c_emb}, the ECM in the test stage can be designed as
\begin{equation}\label{ECM_emb}
X_{ECM} = c_{emb}^{3 \times 3\times 3}(X_{in}).
\end{equation}

Note that although four convolutions are employed to improve the representation ability in the training stage, only a single convolution is required in the test stage, which does not increase the parameter number and not degrade the inference running efficiency.  

\subsection{Multi-Level Framework}
To further improve the representation ability, we design the multi-level framework based on the proposed ECM block for both high-level and low-level feature extraction. The low-level branch directly extracts features from the large-size feature maps,  which focuses on details of the human body. By  contrast, the high-level one which works on down-sampled feature maps based on max pooling can extract more abstract information.
\subsection{Temporal Aggregation and Spatial Mapping}
\label{FP_FM}
To adaptively aggregate the temporal information of variable-length gait sequences, we introduce the temporal aggregation~\cite{lin2020gait}. Assuming that the feature map $X_{out}\in \mathbb{R}^{C_f \times T_f \times H_f \times W_f}$ is the output of the feature extraction module, the temporal aggregation operation can be represented as
\begin{equation}
  Y_{ta} = F_{Max}^{1\times T_f\times1\times1}(X_{out}),
\end{equation}
where $Y_{ta} \in \mathbb{R}^{C_f \times 1 \times H_f \times W_f}$ is the output of the temporal aggregation module.

For the spatial mapping, we generate multiple horizontal feature representations and then combine them to improve the spatial representation ability \cite{lin2020gait,fan2020gaitpart,chao2019gaitset,yu20206dof,liu2020leaping}. The spatial mapping can be represented as
\begin{equation}
  Y_{out} = F_s( F_{GeM}^{1\times1\times1\times W_f}(Y_{ta})  ),
\end{equation}
%+ F_{Avg}^{1\times1\times1\times W_f}(Y_{ta})
where $Y_{out} \in \mathbb{R}^{C_f \times 1\times H_f\times 1}$ is the output of the spatial mapping. $F_{GeM} (\cdot)$ means the Generalized-Mean (GeM) pooling operation ~\cite{radenovic2018fine}. $F_s(\cdot)$ denotes the multiple separate fully connected (FC) layers. After spatial mapping, we obtain the final feature representation $Y$ by concatenating the high-level and low-level feature maps in horizontal axis.

\subsection{Loss Function}

To train the proposed network, we employ the combined loss function which consists of the triplet loss and cross entropy loss. Besides the traditional cross entropy loss used for classification, the triplet loss is also introduced to make the samples from the same ID as close as possible while those from different IDs have larger distance in the feature space. 
The combined loss function is calculated with the obtained the output of spatial mapping, which is  represented as
\begin{equation}
L_{combined} = L_{tri} + L_{cse},
\end{equation}
where the $L_{tri}$ and $L_{cse}$ denote the triplet loss and cross entropy loss, respectively. $L_{tri}$ is defined as 
\begin{equation}
\label{tri}
L_{tri} = {\max}(d(r,s) - d(r,t) + m,0)
\end{equation}
where $r$ and $s$ are samples of the same category, while $r$ and $t$ are samples from different categories. $d(\cdot)$ represents the Euclidean distance between the two samples and  $m$ is the margin of the triplet loss.

\subsection{Training and Test Details}\label{training_test}
\noindent \textbf{Training.} 
In this paper, we introduce a combined loss function consisting of cross-entropy loss and triplet loss~\cite{chao2019gaitset,fan2020gaitpart,yu2019unsupervised,tian2019sosnet,zhang2022gigapixel,zhang2018sleep} to train the proposed GaitStrip. Specifically, the feature representation $Y$ is fed into the triplet loss function for calculation~\cite{chao2019gaitset}, and input into the cross-entropy loss function through an FC layer. The Batch ALL (BA)~\cite{hermans2017defense} is used as the sampling strategy. The number of samples of each batch is $P \times K$, which contains $P$ classes and each class corresponds to $K$ samples. Considering the memory limit, the length of gait sequences is set to $T$ in the training stage.

\noindent \textbf{Test.} 
During the test stage, we input the whole sequences into the GaitStrip to produce the feature representation $Y$. After that, $Y$ is flattened into a vector to represent the corresponding sample. In general, the gait datasets are usually divided into two sets, \ie, the gallery set and the probe set. The feature vectors from the gallery set are taken as the standard view to be retrieved, while those from the probe set are used to evaluate the performance. Specifically, we calculate the Euclidean distance between the feature vectors in the probe set and all feature vectors in the gallery set. The class label of the gallery sample with the smallest distance will be assigned to the probe sample.

\section{Experiments}

\subsection{Datasets and Evaluation Protocol}
\noindent \textbf{CASIA-B.} The CASIA-B dataset~\cite{yu2006framework} is one of the largest gait datasets for evaluation. It includes 124 subjects, each of which contains 10 groups of gait sequences (six groups of normal walking (NM) \#01-\#06, two groups of walking with a bag (BG) \#01-\#02 and two groups of walking with a coat (CL) \#01-\#02). Each group contains 11 view angles ($0^{\circ}$-$180^{\circ}$) and the sampling interval is $18^{\circ}$. Hence, the CASIA-B dataset contains 124 (subject) $\times$ 10 (groups) $\times$ 11 (view angle) = 13,640 gait sequences. The dataset is divided into two subsets, the training set and the test set. We use the protocol~\cite{chao2019gaitset} for evaluation, which includes three different settings, i.e., Small-sample Training(ST), Medium-sample Training (MT) and Large-sample Training (LT). In the three settings, 24, 62 and 74 subjects are used to form the training set, respectively, and the rest are used for test. During the test stage, four groups of sequences (NM\#01-\#04) are used as the gallery set and the rest (NM\#05-\#06, BG\#01-\#02 and CL\#01-\#02) are taken as the probe set.

\noindent \textbf{OUMVLP.} The OUMVLP dataset~\cite{takemura2018multi} is one of the most popular gait datasets, which includes 10,307 subjects. 
Each subject was collected in two groups of video sequences (Seq\#00 and Seq\#01), each of which contains 14 view angles ($0^\circ$, $15^\circ$,...,$75^\circ$, $90^\circ$, $180^\circ$, $195^\circ$,...,$255^\circ$, $270^\circ$). During the test phase, the sequences in Seq\#01 are used as the gallery set and the sequences in Seq\#00 are taken as the probe set.

\begin{table*}[ht]\tiny
%   \scriptsize
  \centering
  %\renewcommand\arraystretch{1.1}
%   \vspace*{-2em}
  \caption{Rank-1 accuracy (\%) on CASIA-B under all view angles, different settings and conditions, excluding identical-view case.
 }%\vspace{-0.5em}
  \resizebox{0.96\textwidth}{!}{
    \begin{tabular}{c|c|c|c|c|c|c|c|c|c|c|c|c|c|c}
    \toprule
    \multicolumn{3}{c|}{Gallery NM\#1-4}  &\multicolumn{12}{c}{$0^{\circ}$-$180^{\circ}$} \\
    \hline
    \multicolumn{3}{c|}{Probe}    & $0^{\circ}$     & $18^{\circ}$    & $36^{\circ}$    & $54^{\circ}$    & $72^{\circ}$    & $90^{\circ}$    & $108^{\circ}$   & $126^{\circ}$   & $144^{\circ}$   & $162^{\circ}$   & $180^{\circ}$  & Mean\\
    \midrule

    \multicolumn{1}{c|}{\multirow{12}[2]{*}{\textbf{ST (24)}}} & \multicolumn{1}{c|}{\multirow{4}[2]{*}{NM\#5-6}} & GaitSet \cite{chao2019gaitset} & 64.6  & 83.3  & 90.4  & 86.5  & 80.2  & 75.5  & 80.3  & 86.0  & 87.1  & 81.4  & 59.6  & 79.5   \\
\cline{3-15}          &       & MT3D \cite{lin2020gait}  & 71.9  & 83.9  & 90.9  & 90.1  & 81.1  & 75.6  & 82.1  & 89.0  & 91.1  & 86.3  & 69.2  & 82.8   \\
\cline{3-15}          &       & GaitGL \cite{lin2021gait} & 77.0  & 87.8  & 93.9  & 92.7  & 83.9  & 78.7  & 84.7  & 91.5  & 92.5  & 89.3  & 74.4  & 86.0   \\
\cline{3-15}          &       & Ours  & \textbf{79.6} & \textbf{89.5} & \textbf{95.6} & \textbf{94.3} & \textbf{86.4} & \textbf{82.0} & \textbf{86.6} & \textbf{93.0} & \textbf{93.6} & \textbf{90.1} & \textbf{75.1} & \textbf{87.8}  \\
\cline{2-15}          & \multicolumn{1}{c|}{\multirow{4}[2]{*}{BG\#1-2}} & GaitSet \cite{chao2019gaitset} & 55.8  & 70.5  & 76.9  & 75.5  & 69.7  & 63.4  & 68.0  & 75.8  & 76.2  & 70.7  & 52.5  & 68.6   \\
\cline{3-15}          &       & MT3D \cite{lin2020gait}  & 64.5  & 76.7  & 82.8  & 82.8  & 73.2  & 66.9  & 74.0  & 81.9  & 84.8  & 80.2  & 63.0  & 74.0   \\
\cline{3-15}          &       & GaitGL \cite{lin2021gait} & 68.1  & 81.2  & 87.7  & 84.9  & 76.3  & 70.5  & 76.1  & 84.5  & 87.0  & 83.6  & 65.0  & 78.6   \\
\cline{3-15}          &       & Ours  & \textbf{71.4} & \textbf{82.6} & \textbf{90.4} & \textbf{88.1} & \textbf{77.9} & \textbf{73.6} & \textbf{79.8} & \textbf{86.4} & \textbf{89.1} & \textbf{86.3} & \textbf{71.3} & \textbf{81.5}  \\
\cline{2-15}          & \multicolumn{1}{c|}{\multirow{4}[2]{*}{CL\#1-2}} & GaitSet \cite{chao2019gaitset} & 29.4  & 43.1  & 49.5  & 48.7  & 42.3  & 40.3  & 44.9  & 47.4  & 43.0  & 35.7  & 25.6  & 40.9   \\
\cline{3-15}          &       & MT3D \cite{lin2020gait}  & 46.6  & 61.6  & 66.5  & 63.3  & 57.4  & 52.1  & 58.1  & 58.9  & 58.5  & 57.4  & 41.9  & 56.6   \\
\cline{3-15}          &       & GaitGL \cite{lin2021gait} & 46.9  & 58.7  & 66.6  & 65.4  & 58.3  & 54.1  & 59.5  & 62.7  & 61.3  & 57.1  & 40.6  & 57.4   \\
\cline{3-15}          &       & Ours  & \textbf{54.3} & \textbf{67.8} & \textbf{75.0} & \textbf{71.6} & \textbf{66.2} & \textbf{59.7} & \textbf{65.5} & \textbf{70.5} & \textbf{69.6} & \textbf{63.6} & \textbf{46.6} & \textbf{64.6}  \\


    \hline
    \multicolumn{1}{c|}{\multirow{12}[2]{*}{\textbf{MT (62)}}} & \multicolumn{1}{c|}{\multirow{4}[2]{*}{NM\#5-6}} & GaitSet \cite{chao2019gaitset} & 86.8  & 95.2  & 98.0  & 94.5  & 91.5  & 89.1  & 91.1  & 95.0  & 97.4  & 93.7  & 80.2  & 92.0   \\
\cline{3-15}          &       & MT3D \cite{lin2020gait}  & 91.9  & 96.4  & 98.5  & 95.7  & 93.8  & 90.8  & 93.9  & 97.3  & 97.9  & 95.0  & 86.8  & 94.4   \\
\cline{3-15}          &       & GaitGL \cite{lin2021gait} & 93.9  & 97.6  & \textbf{98.8} & 97.3  & 95.2  & 92.7  & 95.6  & 98.1  & 98.5  & 96.5  & 91.2  & 95.9   \\
\cline{3-15}          &       & Ours  & \textbf{94.0} & \textbf{98.0} & 98.7  & \textbf{97.8} & \textbf{95.6} & \textbf{93.0} & \textbf{96.1} & \textbf{98.2} & \textbf{98.6} & \textbf{97.0} & \textbf{92.6} & \textbf{96.3}  \\
\cline{2-15}          & \multicolumn{1}{c|}{\multirow{4}[2]{*}{BG\#1-2}} & GaitSet \cite{chao2019gaitset} & 79.9  & 89.8  & 91.2  & 86.7  & 81.6  & 76.7  & 81.0  & 88.2  & 90.3  & 88.5  & 73.0  & 84.3   \\
\cline{3-15}          &       & MT3D \cite{lin2020gait}  & 86.7  & 92.9  & 94.9  & 92.8  & 88.5  & 82.5  & 87.5  & 92.5  & 95.3  & 92.9  & 81.2  & 89.8   \\
\cline{3-15}          &       & GaitGL \cite{lin2021gait} & 88.5  & 95.1  & 95.9  & 94.2  & 91.5  & 85.4  & 89.0  & 95.4  & 97.4  & 94.3  & 86.3  & 92.1   \\
\cline{3-15}          &       & Ours  & \textbf{88.8} & \textbf{95.2} & \textbf{96.8} & \textbf{95.5} & \textbf{92.7} & \textbf{87.4} & \textbf{90.7} & \textbf{95.7} & \textbf{97.6} & \textbf{95.3} & \textbf{87.0} & \textbf{93.0}  \\
\cline{2-15}          & \multicolumn{1}{c|}{\multirow{4}[2]{*}{CL\#1-2}} & GaitSet \cite{chao2019gaitset} & 52.0  & 66.0  & 72.8  & 69.3  & 63.1  & 61.2  & 63.5  & 66.5  & 67.5  & 60.0  & 45.9  & 62.5   \\
\cline{3-15}          &       & MT3D \cite{lin2020gait}  & 67.5  & 81.0  & 85.0  & 80.6  & 75.9  & 69.8  & 76.8  & 81.0  & 80.8  & 73.8  & 59.0  & 75.6   \\
\cline{3-15}          &       & GaitGL \cite{lin2021gait} & 70.7  & 83.2  & 87.1  & 84.7  & 78.2  & 71.3  & 78.0  & 83.7  & 83.6  & 77.1  & 63.1  & 78.3   \\
\cline{3-15}          &       & Ours  & \textbf{69.2} & \textbf{86.7} & \textbf{90.0} & \textbf{88.3} & \textbf{83.6} & \textbf{75.8} & \textbf{82.3} & \textbf{88.1} & \textbf{88.1} & \textbf{81.7} & \textbf{65.7} & \textbf{81.8}  \\
    \hline

    \multicolumn{1}{c|}{\multirow{15}[2]{*}{\textbf{LT (74)}}} & \multicolumn{1}{c|}{\multirow{5}[2]{*}{NM\#5-6}} & GaitSet \cite{chao2019gaitset} & 90.8  & 97.9  & 99.4  & 96.9  & 93.6  & 91.7  & 95.0  & 97.8  & 98.9  & 96.8  & 85.8  & 95.0   \\
\cline{3-15}          &       & GaitPart \cite{fan2020gaitpart} & 94.1  & 98.6  & 99.3  & 98.5  & 94.0  & 92.3  & 95.9  & 98.4  & 99.2  & 97.8  & 90.4  & 96.2   \\
\cline{3-15}          &       & MT3D \cite{lin2020gait}  & 95.7  & 98.2  & 99.0  & 97.5  & 95.1  & 93.9  & 96.1  & 98.6  & 99.2  & 98.2  & 92.0  & 96.7   \\
\cline{3-15}          &       & GaitGL \cite{lin2021gait} & 96.0  & 98.3  & \textbf{99.0} & 97.9  & \textbf{96.9} & \textbf{95.4} & 97.0  & 98.9  & \textbf{99.3} & 98.8  & 94.0  & 97.4   \\
\cline{3-15}          &       & Ours  & \textbf{96.0} & \textbf{98.4} & 98.8  & \textbf{97.9} & 96.6  & 95.3  & \textbf{97.5} & \textbf{98.9} & 99.1  & \textbf{99.0} & \textbf{96.3} & \textbf{97.6}  \\
\cline{2-15}          & \multicolumn{1}{c|}{\multirow{5}[2]{*}{BG\#1-2}} & GaitSet \cite{chao2019gaitset} & 83.8  & 91.2  & 91.8  & 88.8  & 83.3  & 81.0  & 84.1  & 90.0  & 92.2  & 94.4  & 79.0  & 87.2   \\
\cline{3-15}          &       & GaitPart \cite{fan2020gaitpart} & 89.1  & 94.8  & 96.7  & 95.1  & 88.3  & 84.9  & 89.0  & 93.5  & 96.1  & 93.8  & 85.8  & 91.5   \\
\cline{3-15}          &       & MT3D \cite{lin2020gait}  & 91.0  & 95.4  & 97.5  & 94.2  & 92.3  & 86.9  & 91.2  & 95.6  & 97.3  & 96.4  & 86.6  & 93.0   \\
\cline{3-15}          &       & GaitGL \cite{lin2021gait} & 92.6  & 96.6  & 96.8  & 95.5  & 93.5  & 89.3  & 92.2  & 96.5  & 98.2  & 96.9  & \textbf{91.5} & 94.5   \\
\cline{3-15}          &       & Ours  & \textbf{92.8} & \textbf{96.6} & \textbf{97.2} & \textbf{96.5} & \textbf{95.2} & \textbf{90.5} & \textbf{93.5} & \textbf{97.5} & \textbf{98.3} & \textbf{97.6} & 91.4  & \textbf{95.2}  \\
\cline{2-15}          & \multicolumn{1}{c|}{\multirow{5}[2]{*}{CL\#1-2}} & GaitSet \cite{chao2019gaitset} & 61.4  & 75.4  & 80.7  & 77.3  & 72.1  & 70.1  & 71.5  & 73.5  & 73.5  & 68.4  & 50.0  & 70.4   \\
\cline{3-15}          &       & GaitPart \cite{fan2020gaitpart} & 70.7  & 85.5  & 86.9  & 83.3  & 77.1  & 72.5  & 76.9  & 82.2  & 83.8  & 80.2  & 66.5  & 78.7   \\
\cline{3-15}          &       & MT3D \cite{lin2020gait}  & 76.0  & 87.6  & 89.8  & 85.0  & 81.2  & 75.7  & 81.0  & 84.5  & 85.4  & 82.2  & 68.1  & 81.5   \\
\cline{3-15}          &       & GaitGL \cite{lin2021gait} & 76.6  & 90.0  & 90.3  & 87.1  & 84.5  & 79.0  & 84.1  & 87.0  & 87.3  & 84.4  & 69.5  & 83.6   \\
\cline{3-15}          &       & Ours  & \textbf{79.9} & \textbf{92.3} & \textbf{93.4} & \textbf{89.2} & \textbf{86.0} & \textbf{80.0} & \textbf{86.0} & \textbf{88.5} & \textbf{91.7} & \textbf{87.5} & \textbf{73.5} & \textbf{86.2}  \\

    \bottomrule

    \end{tabular}%
  }

\label{comparision_casia}
% \vspace*{-2em}
\end{table*}%

\subsection{Implementation Details}
Gait sequences are preprocessed and normalized into the same size  ${64\times44}$ on both datasets~\cite{chao2019gaitset}. 
% The channel number of the proposed GaitStrip is shown in Table \ref{tab_network}. 
In CASIA-B, GaitStrip has four blocks, where the last three blocks are built with the proposed ECM. The channel number of the four blocks is set to 32, 64, 128 and 128, respectively.
% In CASIA-B, the output channel of the feature extractor is 32, while in the low-level branch ECM's input-channel and output-channel is 32 and 64, in the high-level branch the ECM's input-channel and output-channel is 64 and 128. 
In OUMVLP, we use five blocks to construct the proposed GaitStrip and the last two blocks are implemented by the ECM module. The channel number of the five blocks is set to 64, 128, 196, 256 and 256, respectively.
% In OUMVLP, the output-channel of the feature extractor is 64. The output-channel of the low-level branch is 196, while the output-channel of the high-level is 256.
The margin of the triplet loss is set to 0.2 and Adam is selected as the optimizer.
During the training stage, the parameters $P$ and $K$ are both set to 8. And the length of sequences $T$ is set to 30. The learning rate is set to 1e-4 and reset to 1e-5 in the last 10K iterations. For the settings ST, MT and LT on CASIA-B dataset, the iteration number is set to 60K, 70K and 80K, respectively. On OUMVLP dataset, the parameter $P \times K$ is set to $32 \times 8$. The iteration number is set to 210K. The learning rate is initialized to 1e-4 and reset to 1e-5 after 150K iterations.

\subsection{Comparison with the State-of-the-Art}

\noindent \textbf{Evaluation on CASIA-B.} We compare the proposed method with several gait recognition approaches including GaitSet~\cite{chao2019gaitset}, GaitPart~\cite{fan2020gaitpart}, MT3D~\cite{lin2020gait} and GaitGL~\cite{lin2021gait} on the CASIA-B dataset. The experimental results are shown in Table~\ref{comparision_casia}. It can be observed that the proposed method achieves the highest average accuracy under all settings (ST, MT and LT) and conditions (NM, BG and CL). 
%The results of the experiment achieve the highest performance in different settings(ST, MT and LT) in various conditions(NM, BG and CL) and in some special views. 
Furthermore, we explore the performance of the proposed method under different settings and conditions in details. 

\noindent \textbf{Evaluation under various settings (ST, MT and LT).} We observe that our method achieves high performance under all three settings (ST, MT and LT) and exceeds the best result reported before. We display the complete experimental results under these three settings in Table~\ref{comparision_casia}. The recognition accuracy of GaitGL under ST MT and ST settings in NM condition is 86.0\%, 95.9\% and 97.4\%, respectively. For the proposed method, the gait recognition accuracy is 87.8\%, 96.3\% and 97.6\%, respectively.  Furthermore, our method obtains significant improvement comparing with other methods in all three settings.

% \noindent \textbf{Evaluation under various settings (ST, MT and LT).} We observe that our method achieves high performance under all three settings (ST, MT and LT) and exceeds the best result reported before. We display the complete experimental results under these three settings in Table~\ref{comparision_casia}. The recognition accuracy of GaitGL under ST and MT settings in NM condition is 86.0\% and 95.9\%, respectively. For the proposed method, the gait recognition accuracy is 87.8\% and 96.3\%, respectively. The recognition accuracy of GaitGL under LT setting in NM condition is 97.4\%, while the accuracy of our method achieves 97.6\% in the same condition, which increases by 0.2\%. Furthermore, our method obtains significant improvement comparing with other methods in all three settings. 

\noindent \textbf{Evaluation under various conditions (NM, BG and CL).} It can be seen that when the external environment changes and more challenges exist, the accuracy decreases heavily. Under the LT setting, the accuracy of GaitGL in NM, BG and CL conditions is 97.4\%, 94.5\% and 83.6\%, respectively. Comparing with GaitGL, our results are 0.2\%, 0.7\% and 2.6\% higher, respectively.  Under ST and MT settings, we can also observe that the proposed method owns the best performance. In the ST setting, our method outperforms GaitGL by 1.8\%, 2.9\% and 7.2\% under NM, BG and CL, respectively. In the MT setting, the accuracy of the proposed method is 96.3\%, 93.0\% and 81.8\%, which exceeds GaitGL by 0.4\%, 0.9\% and 3.5\%, respectively.

% The experimental results of our method achieve obvious advantages in complex conditions, indicating that features extracted in the proposed method have better discriminability.

\noindent \textbf{Evaluation on specific angles ($0^\circ$, $90^\circ$, $180^\circ$).} The proposed method shows significant improvement in some extreme view angles ($0^{\circ}$, $90^{\circ}$ and $180^{\circ}$). For example, the average accuracy of MT3D in the setting LT and NM is 96.7\%, but the accuracy corresponding to the three specific view angles are 95.7\%, 93.9\% and 92.0\%, respectively. For the proposed method, the accuracy in the setting LT and NM is 97.6\%, which outperforms MT3D by 0.9\%. And the accuracy corresponding to the specific view angles ($0^{\circ}$, $90^{\circ}$ and $180^{\circ}$) are 96.0\%, 95.3\% and 96.3\%, respectively, which outperforms MT3D by 0.3\%, 1.4\% and 4.3\%, respectively. The main reason may be that the proposed SPB module extracts the feature of each strip, making the proposed ECM obtain more effective feature representation in the specific view angles.

\noindent \textbf{Evaluation on OUMVLP.} 
Compared with the CASIA-B, the OUMVLP dataset contains more subjects. Hereby, we compare GaitStrip with several famous gait recognition methods, including GEINet~\cite{shiraga2016geinet}, GaitSet~\cite{chao2019gaitset}, GaitPart~\cite{fan2020gaitpart}, GLN~\cite{hou2020gait}, SRN+CB~\cite{hou2021set}, GaitGL~\cite{lin2021gait} and 3D Local~\cite{huang20213d} on this dataset. The experimental results are shown in Table~\ref{comparision_oumvlp} which indicates that the proposed method achieves the optimal performance in all conditions. For example, the accuracy of GaitGL with invalid probe is 89.7\%. For the proposed method, the accuracy in the same conditions is 90.5\% , which outperforms GaitGL by 0.8\%. The accuracy of GaitGL excluding invalid probe sequences is 96.2\%, while the accuracy of the proposed method is 97.0\%. 

\begin{table*}[htbp]\tiny
%\vspace*{-2em}
      \caption{Rank-1 accuracy (\%) on OUMVLP dataset under different view angles, excluding identical-view cases. 
    %   The top eight rows and bottom seven rows show the results with and without invalid probe sequences, respectively.
      }
  \centering
  \resizebox{0.99\textwidth}{!}{
    \begin{tabular}{cc|c|c|c|c|c|c|c|c|c|c|c|c|c|c}
    \toprule
    \multirow{2}[2]{*}{\textbf{Method}} & \multicolumn{14}{|c|}{\textbf{Probe View}}                                                            & \multicolumn{1}{c}{\multirow{2}[2]{*}{\textbf{Mean}}}  \\
\cline{2-15}    \multicolumn{1}{c|}{} & $0^{\circ}$ & $15^{\circ}$ & $30^{\circ}$ & $45^{\circ}$ & $60^{\circ}$ & $75^{\circ}$ & $90^{\circ}$ & $180^{\circ}$ & $195^{\circ}$ & $210^{\circ}$ & $225^{\circ}$ & $240^{\circ}$ & $255^{\circ}$ & $270^{\circ}$ &   \\
    \midrule
    \multicolumn{1}{c|}{GEINet \cite{shiraga2016geinet}} & 24.9  & 40.7  & 51.6  & 55.1  & 49.8  & 51.1  & 46.4  & 29.2  & 40.7  & 50.5  & 53.3  & 48.4  & 48.6  & 43.5  & 45.3   \\
    \hline
    \multicolumn{1}{c|}{GaitSet \cite{chao2019gaitset}} & 84.5  & 93.3  & 96.7  & 96.6  & 93.5  & 95.3  & 94.2  & 87.0  & 92.5  & 96.0  & 96.0  & 93.0  & 94.3  & 92.7  & 93.3   \\
    \hline
    \multicolumn{1}{c|}{GaitPart \cite{fan2020gaitpart}} & 88.0  & 94.7  & 97.7  & 97.6  & 95.5  & 96.6  & 96.2  & 90.6  & 94.2  & 97.2  & 97.1  & 95.1  & 96.0  & 95.0  & 95.1  \\
    \hline
    \multicolumn{1}{c|}{GLN \cite{hou2020gait}}   & 89.3  & 95.8  & 97.9  & 97.8  & 96.0  & 96.7  & 96.1  & 90.7  & 95.3  & 97.7 & 97.5 & 95.7  & 96.2  & 95.3  & 95.6   \\
    \hline
    \multicolumn{1}{c|}{SRN+CB \cite{hou2021set}} & 91.2  & 96.5  & 98.3  & 98.4  & 96.3  & 97.3  & 96.8  & 92.3  & 96.3  & 98.1  & 98.1  & 96.0  & 97.0  & 96.2  & 96.4  \\
    \hline
    \multicolumn{1}{c|}{GaitGL \cite{lin2021gait}}  & 90.5  & 96.1  & 98.0  & 98.1  & 97.0  & 97.6  & 97.1  & 94.2  & 94.9  & 97.4  & 97.4  & 95.7  & 96.5  & 95.7  & 96.2  \\
    \hline    
    \multicolumn{1}{c|}{3D Local \cite{huang20213d}}  & $-$  & $-$   & $-$  & $-$   & $-$   & $-$   & $-$   & $-$   & $-$   & $-$  & $-$   & $-$   & $-$   & $-$   & 96.5  \\
    \hline        
    
    \multicolumn{1}{c|}{Ours}  & \textbf{92.8} & \textbf{97.0} & \textbf{98.4} & \textbf{98.5} & \textbf{97.6} & \textbf{98.2} & \textbf{97.8} & \textbf{96.0} & \textbf{96.2} & \textbf{97.8} & \textbf{97.9} & \textbf{96.6} & \textbf{97.3} & \textbf{96.7} & \textbf{97.0} \\

    \bottomrule
    \end{tabular}%
  }

  \label{comparision_oumvlp}%
    %   \vspace*{-2em}
\end{table*}%

\begin{table}[h]\scriptsize
  \centering
%\vspace*{-2em}
  \caption{Rank-1 accuracy (\%) of different ECM blocks.}
    \begin{tabular}{c|c|c|c|c|c}
    
    \toprule
    
    ST  & FL  & SPB   & \multicolumn{1}{c|}{NM} & \multicolumn{1}{c|}{BG} & \multicolumn{1}{c}{CL} \\
    
    \midrule
    
    \checkmark     &       &       & 97.4  & 94.9  & 85.3    \\
    \hline
    \checkmark      &       & \checkmark      & 97.4  & 95.2  & 85.5  \\
    \hline
          & \checkmark      &       & 96.2 & 92.9  & 78.5   \\
    \hline
          & \checkmark      & \checkmark      & 97.2  & 94.9 & 85.2    \\
    \hline
    \checkmark      & \checkmark      &       & 97.4  & 95.2  & 85.9   \\
    \hline
    \checkmark      & \checkmark      & \checkmark      & \textbf{97.6}  & \textbf{95.2}  & \textbf{86.2}   \\
    
    \bottomrule
    
    \end{tabular}%
  \label{tab_ECM}%
%   \vspace*{-2em}
\end{table}%

\subsection{Ablation Study} \label{Ablation_Study} 
In this paper, to obtain effective feature representation, we propose the GaitStrip with ECM block, SPB feature extractor and multi-level framework. Therefore, we design several ablation studies to explore the contribution of the key components.

\noindent \textbf{Analysis of the SPB module.} We propose the novel SPB extractor to extract more discriminative gait features. To explore the contribution of the SPB, we first design three groups of comparative experiments, i.e., only using the ST to compare with the combination of ST and SPB, only using the FL to compare with the combination of FL and SPB, and comparing the combination of ST and FL to the combination of ST, FL and SPB. The experimental results are shown in Table~\ref{tab_ECM}. We can find that the performance of the modules with SPB is improved compared with that without SPB. The accuracy of methods with and without SPB in NM condition is very close, but the methods with SPB in CL condition perform better. Specifically, the accuracy in CL condition by using FL is 78.5\%, while the accuracy in CL condition with the combination of FL and SPB is 85.2\%, which increases by 6.7\%. In the CL condition, the accuracy with the combination of FL, ST and SPB is 86.2\%, which increases 0.3\% compared with that with only the combination of FL and ST. Hence, the SPB can help to extract more comprehensive gait features, which plays an important role in recognition improvement.

\begin{table}[h]\scriptsize
  \centering
  %\vspace*{-2em}
  \caption{Rank-1 accuracy (\%) of different levels.}
    \begin{tabular}{c|c|c|c|c}
    \toprule
    \multicolumn{2}{c|}{Multi-Level Structure} & \multirow{2}[2]{*}{NM} & \multirow{2}[2]{*}{BG} & \multirow{2}[2]{*}{CL}  \\
\cline{1-2}    High-Level & Low-Level &       &       &     \\
    \midrule
    \checkmark      &       & 97.3  & 94.4  & 83.4   \\
    \hline
          & \checkmark      & 97.2  & 94.4  & 84.4  \\
    \hline
    \checkmark      & \checkmark      & \textbf{97.6} & \textbf{95.2} & \textbf{86.2} \\
    \bottomrule
    \end{tabular}%
  \label{TAB_twolevel}%
%   \vspace*{-2em}
\end{table}%

\begin{table}[h]\scriptsize
  \centering
    %\vspace*{-2em}
  \caption{The accuracy (\%) of different strip-based modeling on the CASIA-B dataset.}
    \begin{tabular}{c|c|c|c}
    \toprule
    \multirow{2}[2]{*}{Method} & \multirow{2}[2]{*}{NM} & \multirow{2}[2]{*}{BG} & \multirow{2}[2]{*}{CL} \\
          &       &       &      \\
    \midrule
    baseline+ECM & \textbf{97.6} & \textbf{95.2}  & \textbf{86.2}  \\
    \hline
    baseline+ACB & 96.1  & 92.8  & 79.4   \\
    \hline
    baseline+CCA & 80.4  & 75.1  & 67.6  \\
    \bottomrule
    \end{tabular}%
  \label{tab_strip_modeling}%
    % \vspace*{-2em}
\end{table}%

\noindent \textbf{Analysis of the ECM block.} In this paper, we propose the ECM to generate the discriminative feature representations by taking full advantage of the frame-level and strip-based information. The ECM consists of the ST, FL and SPB. To explore the advantage of the combination of the ST, FL and SPB in robust feature extraction, we design ablation experiments by using only one or two modules. The results of the ablation experiments are shown in Table~\ref{tab_ECM}. In NM condition, the accuracy of the combination of ST and SPB is 97.4\%, the accuracy of the combination of FL and SPB is 97.2\%, and the combination of the ST, FL and SPB is 97.6\%, which increases by 0.2\% and 0.4\%, respectively, compared with the other two modules. The accuracy of the study shows that the combination of the ST, FL and SPB can obtain better accuracy in NM, BG and CL conditions than using only one or two of the modules.

\noindent \textbf{Analysis of multi-level framework.} The proposed GaitStrip works with multiple levels. To investigate the contribution of the low-level and high-level branches, we design the comparison methods with only one branch. The experimental results are shown in Table \ref{TAB_twolevel}, from which we can observe that the accuracy of the methods with only high-level or low-level branch is 97.3\% and 97.2\%, respectively, while the accuracy with both levels is 97.6\%, which achieves 0.3\% and 0.4\% improvement, respectively, demonstrating that the multi-level structure can effectively enhance the representation ability and then improve the recognition performance.

\subsection{Comparison with Other Strip-based Modeling}
In Sec.~\ref{related_sbm}, we introduce two different modules to model the strip-based information. To analyze their performance, we design some experiments by using the Asymmetric Convolution Block (ACB) or Criss-Cross Attention Block to replace the ECM module.
All experiments are built with the LT setting on CASIA-B. The experimental results are shown in Table~\ref{tab_strip_modeling}. It can be observed that the proposed ECM achieves better performance than other strip-based modelings. This may be because our ECM utilizes the spatial-temporal information of each strip, improving the feature representation ability.
The accuracy of the ECM method in NM, BG and CL is 97.6\%, 95.2\% and 86.2\% respectively, which exceeds the ACB method by 1.5\%, 2.4\% and 6.8\%. The accuracy of the CCA method in NM, BG and CL is 80.4\%, 75.1\% and 67.7\% respectively, which is inferior to our method as well. 
By comparing with other strip-based methods, we can note that the proposed method can better exploit the spatial-temporal representation, especially in some complex conditions, which achieves significant improvement. 


\subsection{Computational Analysis}

In the inference stage, the proposed ECM can be embedded into a standard 3D convolution, which reduces parameters and inference time.
The computational analysis is shown in Table~\ref{tab_details}. It can be observed that the average accuracy of using ECM is 93.0\%, outperforming the accuracy of using ST by 0.5\%. However, the parameters of both modules are equal.

% By using re-parameterization, ECM improves the accuracy without increasing parameters and inference time.


\begin{table}[htbp]\scriptsize
  \centering
  %\vspace*{-2em}
  \caption{The accuracy(\%), inference time (second/sequence) and parameters (M) of different methods on CASIA-B dataset}
%   \vspace*{-1em}
    \resizebox{0.5\textwidth}{!}{
    \begin{tabular}{lcccc}
    \toprule
          & \multicolumn{1}{l}{Re-param} & ST    & ST+FL & ECM \\
    \midrule
    \hline
    Accuracy &    $-$   & 92.5  & 92.8  & \textbf{93.0}  \\
    \hline
    Inference time  &  $\times$     & 0.025 &   0.027    & 0.035 \\
    \hline
    Parameters &   $\times$    & 3.87 & 4.33 & 5.25 \\
    \hline
    \hline  
    Accuracy & $-$      & 92.5  & 92.8  & \textbf{93.0}  \\
    \hline
    Inference time & \checkmark     & 0.025 & 0.025 & 0.025 \\
    \hline
    Parameters & \checkmark     & 3.87 & 3.87 & 3.87 \\
    \bottomrule
    \end{tabular}
    }
    % \vspace*{-2em}
  \label{tab_details}%
\end{table}%

% \subsection{Visualization}
% To better understand the effectiveness of these branches, we visualize the heatmap figure of each branch, which is shown in Fig.~\ref{fig_visualization}. It can be observed that compared with ST, FL pays more attention to the arms and SPB focuses on the foot movements.

% \begin{figure}[ht]
% \centering
% %\vspace{-1em}
% \includegraphics[width=0.7\textwidth]{1.pdf}\vspace{-1em}
% % \vspace{-0.5em}
% \caption{Visualization of the original sequence and heatmaps for the ST, FL and SPB module.}
% %\vspace{-3em}
% \label{fig_visualization}
% \end{figure}

\section{Conclusion}
In this paper, we propose a novel gait recognition network GaitStrip with ECM block and multi-level framework. 
On the one hand, the proposed ECM which aggregates spatial-temporal, frame-level and strip-based information can generate more comprehensive feature representations. Moreover, the spatial-temporal, frame-level and strip-based feature extractors can be embedded into a common 3D convolution in the inference stage, which does not introduce additional parameters.
On the other hand, the multi-level structure containing both low-level and high-level branches can ensemble global semantic and local detailed information.
The experiment results verify that the proposed GaitStrip achieves appealing performance in normal environment as well as complex conditions.

\noindent \textbf{Acknowledgements.}
% \section{Acknowledgements}
This work was supported by the National Natural Science Foundation of China (61976017 and 61601021), the Beijing Natural Science Foundation (4202056), the Fundamental Research Funds for the Central Universities (2022JBMC013) and the Australian Research Council (DP220100800, DE230100477). The support and resources from the Center for High Performance Computing at Beijing Jiaotong University (http://hpc.bjtu.edu.cn) are gratefully acknowledged.

%===========================================================
%\bibliographystyle{splncs}
\bibliographystyle{splncs04}
\bibliography{egbib}

%this would normally be the end of your paper, but you may also have an appendix
%within the given limit of number of pages
\end{document}